\documentclass[runningheads]{llncs}
\usepackage[T1]{fontenc}
\usepackage{amssymb}
\usepackage{graphicx}
\usepackage{comment}
\usepackage{multirow}
\usepackage{array}
\usepackage{tikz}
\usepackage{enumitem,kantlipsum}
\usepackage{amsmath}
\usepackage{makecell}
\usetikzlibrary{shapes.geometric, arrows.meta, positioning}
\renewcommand{\arraystretch}{1.0}
\newcolumntype{P}[1]{>{\centering\arraybackslash}p{#1}}
\newcolumntype{C}[1]{>{\centering\let\newline\\\arraybackslash\hspace{0pt}}m{#1}}
\let\llncssubparagraph\subparagraph
\let\subparagraph\paragraph
\usepackage[compact]{titlesec}
\let\subparagraph\llncssubparagraph
\begin{document}
\title{From Individual to Group: Developing a Context-Aware Multi-Criteria Group Recommender System}
\titlerunning{Context-Aware Multi-Criteria Group Recommender Systems}
%
%
\author{LE Ngoc Luyen, Marie-Hélène ABEL
}
\institute{Université de technologie de Compiègne, CNRS, Heudiasyc (Heuristics and\\ Diagnosis of Complex Systems), CS 60319 - 60203 Compiègne Cedex, France}
\authorrunning{NL LE et al.}
%
%
\maketitle              
\begin{abstract}
Group decision-making is becoming increasingly common in areas such as education, dining, travel, and finance, where collaborative choices must balance diverse individual preferences. While conventional recommender systems are effective in personalization, they fall short in group settings due to their inability to manage conflicting preferences, contextual factors, and multiple evaluation criteria. This study presents the development of a Context-Aware Multi-Criteria Group Recommender System (CA-MCGRS) designed to address these challenges by integrating contextual factors and multiple criteria to enhance recommendation accuracy. By leveraging a Multi-Head Attention mechanism, our model dynamically weighs the importance of different features. Experiments conducted on an educational dataset with varied ratings and contextual variables demonstrate that CA-MCGRS consistently outperforms other approaches across four scenarios. Our findings underscore the importance of incorporating context and multi-criteria evaluations to improve group recommendations, offering valuable insights for developing more effective group recommender systems.

\end{abstract}
\section{Introduction}

In today's digital age, recommender systems have become indispensable for providing personalized suggestions in various domains such as entertainment, retail, and education \cite{rosa2015music,le2021towards,10152701}. These systems analyze user preferences to suggest items such as movies, products, or learning materials tailored to an individual’s needs. However, many real-world decisions are made collectively, requiring systems that can accommodate the preferences of multiple individuals. This shift from individual to group-based recommendations introduces a new layer of complexity: how to balance the preferences of multiple users to arrive at recommendations that satisfy the group as a whole \cite{masthoff2010group,nozari2020novel}.

Group Recommender Systems (GRS) are designed to tackle this challenge by providing recommendations that reflect the collective preferences of a group. While GRS has made significant strides in various domains, these systems often struggle to address the diverse and sometimes conflicting preferences within groups \cite{dara2020survey}. Additionally, many GRS fail to take into account the multiple criteria on which users may base their decisions, such as quality, price, ease of use, or relevance, as well as the context in which recommendations are made. Multi-criteria factors are important because individuals within the group may prioritize different aspects of the items being recommended.
Contextual factors such as the setting, time, or purpose of the recommendation also play a vital role in group decision-making but are often overlooked in traditional systems \cite{adomavicius2010context,dridi2022exploiting}. This is where Context-Aware Multi-Criteria Group Recommender Systems (CA-MCGRS) come into play, as they consider not only the group’s collective preferences but also evaluate items based on multiple decision criteria while incorporating the context in which the recommendation occurs.

Consider, for example, an educational setting where students work in groups to choose a final project topic. Each student may have different preferences regarding the project’s ease of execution, relevance to their future career, or the quality of data available. Moreover, the context of the decision -- such as whether the project is conducted during a semester with a pandemic lockdown or whether certain resources are available -- can significantly influence the group’s choice. A traditional GRS might focus solely on preferences without accounting for these contextual nuances, leading to suboptimal recommendations \cite{zheng2019integrating}. In contrast, a CA-MCGRS would integrate preferences, context, and multiple criteria to arrive at a project recommendation that satisfies the entire group.

In this paper, we shift the focus from individual to group recommendations and develop Context-Aware Multi-Criteria Group Recommender Systems. We explore how integrating group preferences with contextual data and multi-criteria evaluations can lead to more effective recommendations. Our experiments, conducted using an educational dataset that includes a rich set of multi-criteria ratings (e.g., application relevance, data quality, ease) and contextual factors (e.g., class, semester, lockdown), demonstrate the effectiveness of our approach. The results offer key insights applicable to both academic research and industry settings, where effective group decision-making plays a pivotal role.

The remainder of this paper is organized as follows: In section \ref{section_relatedwork}, we review related work on CA-MCGRSs. Following this, Section \ref{proposition} outlines our primary contributions, including task formulation and the architcture of a CA-MCGRSs. In section \ref{experiments}, we present experimental results of our approach. Finally, we conclude the paper in the last section.

\section{Related Work}\label{section_relatedwork}
In this section, we examine the key developments in group recommender systems, emphasizing the role of context-awareness and multi-criteria approaches in addressing the needs of groups with diverse members.
\subsection{Group Recommender Systems}
Recommender systems (RSs) have become integral in assisting users to navigate vast amounts of information by providing personalized suggestions tailored to individual preferences \cite{resnick1997recommender,luyen_personalized}. Traditionally, these systems focus on enhancing the user experience by predicting and presenting items that align with a single user’s interests \cite{le2023improving}. However, the increasing prevalence of collaborative environments and shared decision-making has spurred interest in GRSs, which aim to aggregate and reconcile the preferences of multiple users to generate recommendations that satisfy the group as a whole \cite{adomavicius2005toward}.

GPSs extend individual recommendation paradigms to accommodate multiple users, addressing unique challenges such as preference aggregation, conflict resolution, and ensuring fairness among group members \cite{xiao2017fairness}. Groups are typically classified as homogeneous, where members share similar interests, or heterogeneous, comprising members with diverse interests \cite{dara2020survey}. Most existing strategies are tailored for homogeneous groups and struggle with heterogeneous ones due to the difficulty in building consensus among differing preferences \cite{kim2010group,gorla2013probabilistic}. Only a limited number of studies, including \cite{sotelo2009tv,quijano2010personality}, have specifically focused on heterogeneous groups.

Current approaches for GRSs can be categorized into \cite{dara2020survey}:
(i) Aggregating Individual Profiles: Combining individual preferences to form a group profile and recommending items based on this collective profile \cite{kim2010group}.
(ii) Aggregating Personalized Recommendations: Generating personalized recommendations for each member and then merging them into a single group recommendation \cite{nozari2020novel}.
Despite these methods, many systems overlook contextual factors that influence group decisions. The incorporation of contextual elements represents a significant area for improvement in the field of GRSs, as they can substantially impact the relevance and effectiveness of recommendations for diverse group scenarios.
\subsection{Integration of Context and Multi-Criteria in Group Recommender Systems}
Context-aware recommender systems enhance traditional models by integrating contextual information into the recommendation process \cite{adomavicius2010context}. By leveraging contextual information, these systems can deliver more relevant and timely recommendations that adapt to the needs and situations of users \cite{abbas2015survey}. Techniques such as contextual filtering, context modeling, and the use of context-aware factorization machines have been employed to effectively incorporate contextual variables \cite{meng2022survey}.

Multi-criteria recommender systems consider multiple attributes or criteria when evaluating and recommending items, providing a more nuanced and comprehensive assessment compared to single-criterion models \cite{manouselis2007analysis}. These systems utilize various methods, including multi-attribute utility theory, weighted sum models, and multi-objective optimization, to balance different criteria \cite{hdioud2017multi}. By accommodating diverse user preferences across multiple dimensions, multi-criteria recommender systems can enhance the relevance and satisfaction of recommendations. Nevertheless, integrating multiple criteria into group settings introduces additional complexity, as it necessitates sophisticated aggregation techniques to balance conflicting criteria preferences among group members.

The integration of context-aware and multi-criteria approaches within group recommender systems represents a promising yet underexplored research area. Recent studies have begun to address this integration by proposing models that simultaneously consider contextual factors and multiple criteria to better capture the complexity of individual preferences \cite{dridi2022exploiting,vu2023deep}. These approaches demonstrate improved recommendation accuracy and individual satisfaction by accounting for situational variables and diverse evaluation metrics. However, their application to group settings remains limited, as most existing models primarily focus on aggregating individual preferences without fully addressing the unique challenges posed by groups. Addressing this gap can enhance the ability of group recommender systems to provide more accurate, relevant, and satisfying recommendations that respect individual preferences within diverse group settings. In the following section, we present our primary approach for developing a context-aware multi-criteria group recommender system, aiming to address the identified gaps by seamlessly integrating contextual information with multiple criteria tailored for groups.

\section{Developing Context-Aware Multi-Criteria Group Recommender Systems}\label{proposition}
In this section, we formalize the problem of designing a CA-MCGRS that incorporates multiple criteria, contextual information, and group dynamics to predict the overall group satisfaction for non-interacted items. We then present our deep neural network architecture for addressing this complex task.

\subsection{Task Formulation}
The objective of developing a CA-MCGRS is to accurately predict and recommend a list of relevant items to groups by integrating contextual factors, multiple evaluation criteria, and considering group size. Therefore, we need to ensure that the recommendations are tailored to collective preferences, contextual influences, and various criteria.

Given that users $\ U = \{u_1, u_2, \dots, u_n\}$ represents the set of individual users; groups  $G = \{g_1, g_2, \dots, g_k\}$ denotes  the set of groups, where each group $g \in G$ is a subset of $U$, i.e., $g \subseteq U$. Each group $g$ has a size $|g|$, representing the number of members; items $I=\{i_1, i_2, ..., i_m\}$ represents the set of items to be recommended; contexts  $C=\{c_1, c_2, ..., c_p\}$ represents the set of contexts affecting interactions with items, $CR=\{cr_1, cr_2, ..., cr_o\}$ represents the set of criteria through ratings,  and $R=\{r_1, r_2, ..., r_q\}$ represents the set of overall ratings. The task recommendation of CA-MCGRS can be defined as follows:
\begin{equation}
	f(r) : G \times I \times C  \times CR \longrightarrow R 
\end{equation} where $G \times I \times C \times CR$ represents the Cartesian product of groups, items, contexts, and criteria, covering all possible group-item interactions influenced by different contextual factors and criteria. Each group \( g \in G \) is a collection of several users. \( CR \) encompasses multiple criteria ratings, providing an evaluation of items depending on various aspects. Each criteria rating \( cr \) is given by group \( g \) to item \( i \) under context \( c \). The overall rating \( R \) is derived from the evaluation of the group for an item under a given context. This overall rating value serves as the primary metric for generating recommendations in the CA-MCGRS.

For each group \( g \in G \), the goal of the CA-MCGRS is to recommend a ranked list of the top-K items \( I_{topK}^g \subseteq I_{ni}^g \) that are most likely to receive high overall ratings \( R(g, i, c) \). This ranking process should take into account the following factors: (i) Contextual Factors: the contextual elements that influence how the group interacts with the items. (ii) Multi-Criteria Evaluations: assessments based on multiple criteria that evaluate different aspects of each item.

	
	

The primary objective is to maximize the overall ratings of the recommended items. This optimization goal can be mathematically expressed as:

\begin{equation}
\max_{I_{topK}^g} \sum_{i \in I_{topK}^g} r(g, i, c, cr)
\end{equation} 
where item $I_{ni}^g$ denotes the set of items that group $g$ has not yet interacted with.
	And $r(g, i, c, cr)$ is the overall rating assigned to item $i$ by group $g$ under context $c$ based on criteria $cr$.

To illustrate how recommendations are made for groups within a specific context and across multiple criteria, consider the following scenario where a university class is selecting projects for different student groups:

\begin{itemize}
	\item \textbf{Students}: \( U = \{u_1, u_2, u_3, u_4, u_5, u_6, u_7, u_8, u_9, u_{10}, u_{11}, u_{12}\} \)
	\item \textbf{Groups}:  \( g_1 = \{u_1, u_2, u_3\} \), ($|g_1| = 3$); \( g_2 = \{u_4, u_5\} \), ($|g_2| = 2$); \( g_3 = \{u_6, u_7,u_8, u_9\} \), ($|g_3| = 4$); \( g_4 = \{u_{10}, u_{11}, u_{12}\} \), ($|g_4| = 3$)
	
	\item \textbf{Projects}: $I = \{$\textit{File Management System}, \textit{Question Answering System}, \textit{Mushroom Classification}, \textit{Zika Virus Epidemic}$\}$
	
	\item \textbf{Contexts}: \( C = \{\text{Class (}c_1\text{)}, \text{Semester (}c_2\text{)}, \text{Lockdown (}c_3\text{)}\} \)
	\item \textbf{Criteria}: \( CR = \{\text{Applicability (}cr_1\text{)}, \text{Data Quality (}cr_2\text{)}, \text{Ease of Use (}cr_3\text{)}\} \)
	\end{itemize}
	
	\vspace{-0.7cm}
	\begin{table}[h!]
		\centering
		\caption{Groups' Evaluation of Projects under Various Contexts and Criteria}
		\label{tab:vehicle_ratings}
		\begin{tabular}{|C{0.9cm}|C{4.3cm}|C{1.0cm}|C{1.5cm}|C{1.0cm}|C{0.6cm}|C{0.6cm}|C{0.6cm}|C{0.6cm}|}
			\hline
			\multirow{2}{*}{\( G \)} & \multirow{2}{*}{\( I \)} & \multicolumn{3}{c|}{\textbf{\( C \)}} & \multicolumn{3}{c|}{\textbf{\( CR \)}} & \multirow{2}{*}{\( R \)} \\\cline{3-8}
			& & \( c_1 \) & \( c_2 \) & \( c_3 \) & \( cr_1 \) & \( cr_2 \) & \( cr_3 \) & \\ \hline
			$g_1$ & File Management System & DM & Spring & POS & 5 & 5 & 4 & 5 \\ \hline
			$g_2$ & Question Answering system & DA & Fall & POS & 4 & 4 & 4 & 3 \\ \hline
			$g_3$ & Mushroom Classification & DB & Spring & PRE & 3 & 5 & 4 & 3 \\ \hline
			$g_4$ & Zika Virus Epidemic & DM & Spring & PRE & 2 & 4 & 5 & 5 \\ \hline
		\end{tabular}
	\end{table}	
		\vspace{-0.5cm}
		
In this scenario, the RS needs to account for group size and composition when aggregating preferences. It must also integrate contextual and multi-criteria evaluations to recommend projects that align with the group's collective preferences. For example, recommending the ``\textit{File Management System}'' project to $g_1$ (with group size: 3) involves considering how the group’s preferences, contextual factors, and criteria ratings predict the best project match for that group.
	
The task formulation for a CA-MCGRS involves defining the sets of users, groups (including their sizes), items, contexts, and criteria, as well as establishing a function that maps these elements to overall ratings. The primary objective is to predict and recommend the top-K items to each group by leveraging multi-criteria ratings and contextual information. In the following section, we will delve into our deep neural network architecture, which is designed to effectively perform such recommendations.

\subsection{Our Deep Neural Network Architecture for CA-MCGRS Leveraging Multi-Head Attention Mechanism}
\begin{figure}[h!]
	\vspace{-0.6cm}
	\centering
	\includegraphics[width=0.9\textwidth]{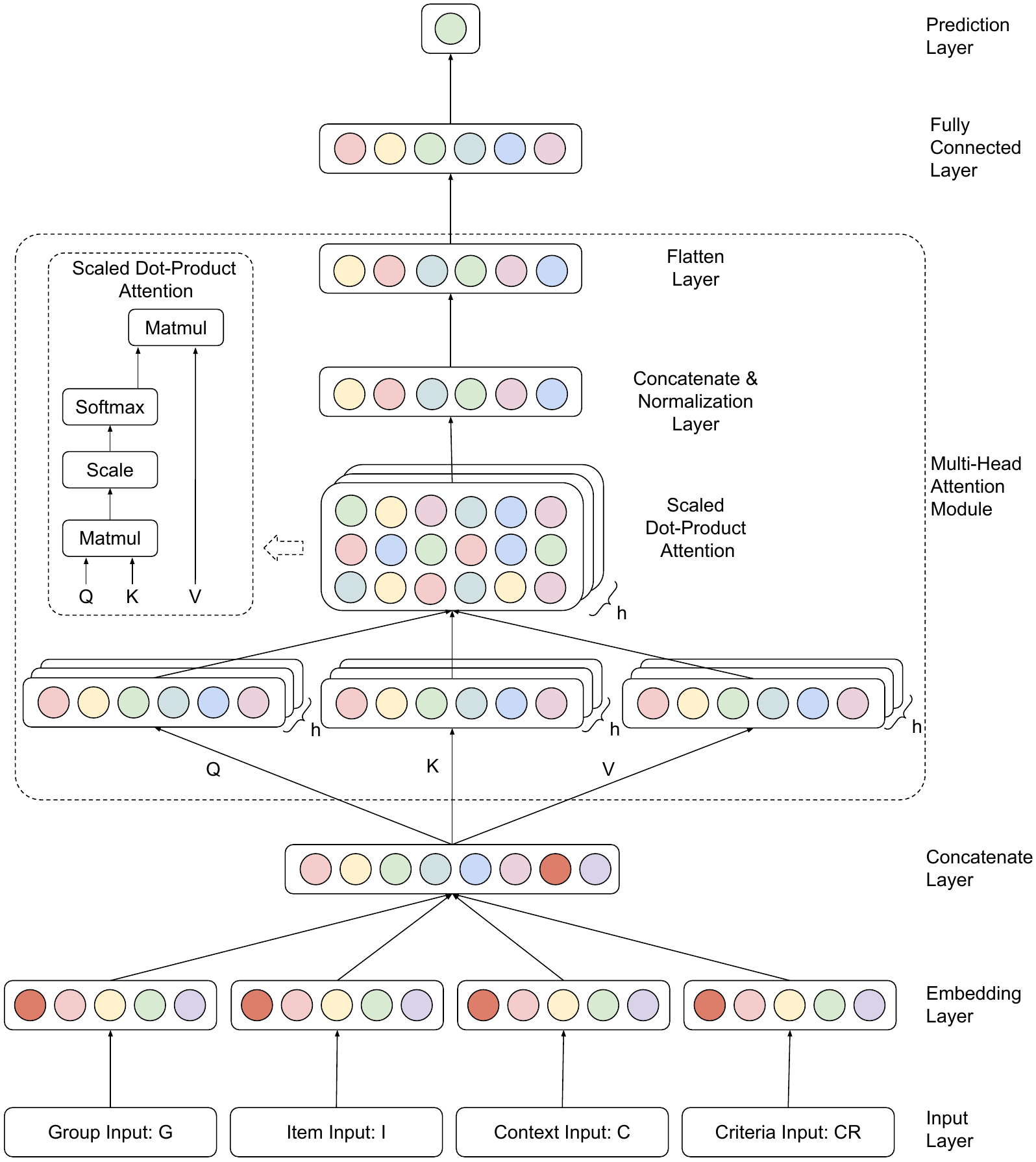}	
	\caption{The deep neural network architecture for the CA-MCGRS using Multi-Head Attention Mechanism.} \label{fig1}
	
	\vspace{-0.6cm}
\end{figure}

In this section, we introduce the CA-MCGRS architecture, which builds upon a dynamic feature selection process. This architecture integrates group preferences along with contextual inputs such as criteria and item information. The model employs multi-head attention mechanism to extract the most relevant features for each recommendation task. The deep neural network architecture consists of several key components that contribute to its adaptability and performance, as illustrated in Figure~\ref{fig1}.

\subsubsection{Input and Embedding Layer}
The input of the architecture includes group preferences, item attributes, contextual inputs, and criteria. Each input then is tranformed and represented as a high-dimensional feature vector, with both sparse (categorical) and dense (continuous) features. Sparse features are converted into lower-dimensional embeddings, while dense features are normalized and used directly. Conventionally, the embeddings for the various inputs are represented as follows:  \textit{$E_G$ = Embedding($G$)} for group inputs, \textit{$E_I$=Embedding($I$)} for item inputs, \textit{$E_C$=Embedding($C$)} for context inputs, and 
\textit{$E_{CR}$=Embedding($CR$)} for criteria inputs.
\subsubsection{Concatenate Layer}: Following the input and embedding layers, the concatenated layer is formed by combining the embeddings from different embeddings:
\begin{equation}
E_{\text{cc}} = [E_G, E_I, E_C, E_{CR}]	
\end{equation}
where $E_G$, $E_I$, $E_C$, and $E_{CR}$ represent the embeddings for group, item, context, and criteria inputs respectively. This concatenation creates a unified feature vector for further processing in the neural network.

\subsubsection{Multi-Head Attention Module}
At the core of the architecture is the multi-head attention mechanism, first introduced in the ``\textit{Attention is All You Need}'' paper by \cite{vaswani2017attention}, which laid the foundation for advanced models such as transformers, widely used in large language models and generative AI. The multi-head attention module operates on embedded feature representations from each input, dynamically identifying the most relevant features by computing multiple attention heads. Each head focuses on different aspects of the input data, capturing complex interactions between groups, items, context, and criteria. Each head computes scaled dot-product attention as follows:

\begin{equation}
\text{Attention}(Q, K, V) = \text{Softmax}\left(\frac{QK^\top}{\sqrt{d}}\right) V
\end{equation}

For each attention head, the query \( Q \), key \( K \), and value \( V \) are projections of the $E_{\text{cc}} $ embeddings:
\begin{equation}
Q = W_Q E_{\text{cc}}, \quad K = W_K E_{\text{cc}}, \quad V = W_V E_{\text{cc}}	
\end{equation}
where \( W_Q, W_K, W_V \in \mathbb{R}^{d \times d_h} \), and \( d_h \) is the dimension of each attention head. The outputs from different heads are concatenated and linearly transformed:

\begin{equation}
Z = \text{MultiHead}(Q,K,V) = [\text{head}_1, \text{head}_2, \dots, \text{head}_h]W_O 
\end{equation}
where \( h \) is the number of heads, and \( W_O \in \mathbb{R}^{d \times d} \) is a projection matrix. Each attention head focuses on different interactions between the features, with each head computed as: 
\begin{equation}
	head_i = \text{Attention}(W^i_Q E_{\text{cc}}, W^i_K E_{\text{cc}}, W^i_V E_{\text{cc}})
\end{equation}

The step where the output of each attention head is concatenated and passed through a normalization layer ensures stability and improves the training process. The concatenation combines information from multiple heads that each focused on different feature interactions. After concatenation, layer normalization is applied to standardize the distribution of the outputs, preventing internal covariate shifts \cite{lei2016layer}. Conventionally, the normalization layer is defined as:
\begin{equation}
Z_{LN} = \frac{Z - \mu}{\sigma}
\end{equation}
where $Z$ is the concatenated output from multiple attention heads, $\mu$ is the mean of $Z$, and $\sigma$ is the standard deviation of $Z$.

Following the normalization, a Flatten Layer converts the multi-dimensional tensor output $Z_{LN}$ into a 1D vector $Z_{Flat}$. This prepares the data for fully connected layers by reshaping it while retaining the information from previous layers.

\subsubsection{Fully Connected Layer}
The output from the flatten layer, \( Z_{\text{flat}} \), is passed into a fully connected dense layer, which captures higher-order interactions between the flattened features. The dense layer applies a weight matrix and bias, followed by a non-linear activation function:

\begin{equation}
H_{\text{dense}} = \text{ReLU}(W_{\text{dense}} Z_{\text{flat}} + b_{\text{dense}})
\end{equation}
where \( W_{\text{dense}} \) represents the weights, \( b_{\text{dense}} \) is the bias, and ReLU introduces non-linearity.

\subsubsection{Prediction Layer}
The final layer produces the model's output by applying a linear transformation to the fully connected layer's output, projecting it to the desired number of output units as follows:

\begin{equation}
\hat{R} = W_{\text{out}} H_{\text{dense}} + b_{\text{out}}
\end{equation}
where \( W_{\text{out}} \) represents the output weights, \( b_{\text{out}} \) is the output bias, and \( \hat{R} \) is the predicted score or rating. The objective of this layer is to minimize the difference between the predicted and actual values by utilizing a regression loss function.

\subsection{Loss Function and Optimization}
In our training process, we use the Mean Squared Error (\textit{MSE}) as the loss function, which is mathematically defined as:

\begin{equation}
L(\theta) = \frac{1}{N} \sum_{i=1}^{N} \left( \hat{R}_i - R_i \right)^2
\end{equation}
where \( N \) is the total number of historical interaction data,\( \hat{R}_i \) is the predicted value for the \(i\)-th sample, \( R_i \) is the true value for the \(i\)-th sample.

For the optimization algorithm, we used is \textit{Adagrad}, which adapts the learning rate for each parameter individually based on the historical gradient information \cite{duchi2011adaptive}. The update rule for \textit{Adagrad} is given by:

\begin{equation}
\theta_{t+1} = \theta_t - \frac{\eta}{\sqrt{S_t + \epsilon}} \cdot g_t
\end{equation}
where \( \theta_t \) is the parameter at time step \( t \), \( \eta \) is the global learning rate, \( S_t \) is the sum of squares of past gradients up to time \( t \), \( \epsilon \) is a small constant added to avoid division by zero, \( s_t \) is the gradient of the loss function at time \( t \).

Overall, our architecture allows CA-MCGRS to dynamically adapt to different contexts and criteria, ensuring that recommendations are both highly relevant and precise by leveraging the multi-head attention mechanism. In the next section, we present experiments designed to assess its performance.

\section{Experiments}\label{experiments}
In this section, we introduce the dataset utilized to assess the performance of our approach and describe the baseline models used for comparison. Finally, we provide a detailed analysis of the experimental results.

\subsection{Dataset}
We conducted our experiments using the ITM-Rec dataset \cite{zheng2023itm}. This dataset is specifically tailored for both group-based and individual-based recommendation tasks in educational contexts. It was collected from individual and group interactions and evaluations in the ITM department of Illinois Institute of Technology, USA. The dataset includes individual and group ratings based on various criteria and contextual factors, allowing for the development and evaluation of the CA-MCGRS. Table \ref{tab1} provides a summary of the key statistics and characteristics of the individuals and groups in the dataset.
\begin{table}[h]
	\vspace{-0.3cm}
	\renewcommand{\arraystretch}{1.6}
	\caption{Statistics on the ITM-Rec dataset}\label{tab1}
	\begin{tabular}{|C{1.7cm}|C{1.4cm}|C{1.0cm}|C{1.8cm}|C{1.4cm}|C{1.2cm}|C{1.4cm}|C{1.4cm}|}
		\hline
		Object &  Quantity & Nb of Item & Contexts & Criteria & Rating Scale & Data Sparsity & Nb of Rating \\
		\hline
		Individual & 454 &\multirow{2}{*}{ 70} & \multirow{2}{*}{\makecell{Class, \\Semester, \\Lockdown}} & \multirow{2}{*}{\makecell{App, \\Data,\\ Ease}} & \multirow{2}{*}{[1,5]} &  83.54\% & 5230\\ \cline{1-2}\cline{7-8}
		Group & 143 & & &  & &  88.84\% & 1117\\\hline
		
	\end{tabular}
\end{table}
\vspace{-0.6cm}

\subsection{Baseline models}
To evaluate our model's effectiveness, we compare its performance with several established baseline methods:
\begin{itemize}[wide, labelwidth=!, labelindent=0pt]
	\item \textbf{AutoInt} learns feature interactions by projecting features into latent spaces, where attention mechanisms identify important interactions \cite{song2019autoint}. \item \textbf{DCN} Deep Cross Network captures feature interactions through a cross network structure, modeling relationships at multiple representation levels \cite{wang2017deep}. \item \textbf{DeepFM} captures complex feature interactions using a deep neural network, leveraging nonlinear activations to model intricate patterns \cite{guo2017deepfm}. \item \textbf{FiBiNET} combines feature importance and bilinear interactions, using a Squeeze-Excitation network to learn feature importance \cite{huang2019fibinet}. \item \textbf{NFM} combines Factorization Machines (FM) with neural networks, FM for basic interactions and neural networks for higher-order feature relationships \cite{he2017neural}.
	\item \textbf{ONN} Operation-aware Neural Networks enhance user response prediction by explicitly modeling feature interactions using predefined operations \cite{yang2020operation}.
	\item \textbf{WDL} captures feature interactions by incorporating explicit interaction terms in a wide linear layer \cite{cheng2016wide}.
	 \item \textbf{xDeepFM} integrates a Compressed Interaction Network for explicit vector-wise interactions, learning both bounded and unbounded interactions \cite{lian2018xdeepfm}.
\end{itemize}
The experiments were carried out using the PyTorch framework \cite{paszke2019pytorch} and the DeepCTR-Torch library \cite{shen2017deepctr}. We relied on the default hyperparameters for each model, as they have been pre-optimized for performance.  the number of heads for the Multi-head Attention module was set to 4, ensuring the model captures various aspects of the input data through multiple attention mechanisms. 
\subsection{Evaluation Metrics}
 We evaluate the deep neural architecture using two standard metrics: Mean Absolute Error (MAE) and Root Mean Squared Error (RMSE), both of which measure the difference between predicted and actual values \cite{shani2011evaluating}.

\begin{equation}
	\text{MAE} = \frac{1}{n} \sum_{i=1}^{n} \left| R_i - \hat{R}_i \right|, 
\end{equation}

\begin{equation}
	\text{RMSE} = \sqrt{\frac{1}{n} \sum_{i=1}^{n} \left( R_i - \hat{R}_i \right)^2}
\end{equation}
where \( n \) is the number of observations, \( R_i \) the actual value, and \( \hat{R}_i \) the predicted value. MAE measures the average error, while RMSE penalizes larger errors. Both metrics indicate prediction accuracy, with lower values signifying better performance.
In the following sections, we present the results of our experiments, comparing our model against baseline approaches using these metrics to highlight its advantages.
\subsection{Experiments Results}
In our experimental results, the CA-MCGRS, developed on the Multi-Head Attention mechanism (referred to as MHA for short), consistently outperformed the baseline models across all metrics and scenarios. Specifically, the table \ref{table2} presents results for four scenarios: Group Recommender Systems (GRS) without contextual or criteria-based information, Multi-Criteria Group Recommender Systems (MCGRS) without context, MCGRS (MC) with multiple contextual factors, and MCGRS (SC) with single contextual factors (class context in this case).

\begin{table}[h!]\label{table2}
	
	\centering
	\caption{Final Results Comparison}
	\begin{tabular}{|C{1.7cm}|C{1.3cm}|C{1.2cm}|C{1.2cm}|C{1.2cm}|C{1.2cm}|C{1.2cm}|C{1.2cm}|C{1.2cm}|}
		\hline
		\multirow{2}{*}{\textbf{Model}} & \multicolumn{2}{c|}{\textbf{GRS}} & \multicolumn{2}{c|}{\textbf{MCGRS}} & \multicolumn{2}{c|}{\textbf{\makecell{MCGRS \\ (MC)}}} & \multicolumn{2}{c|}{\textbf{\makecell{MCGRS  \\ (SC)}}} \\ \cline{2-9}
		& \textbf{RMSE} & \textbf{MAE} & \textbf{RMSE} & \textbf{MAE} & \textbf{RMSE} & \textbf{MAE} & \textbf{RMSE} & \textbf{MAE}  \\ \hline
		AutoInt & 1.3765 & 1.1457 & 1.0123 & 0.8254 & 0.9907 & 0.8023 & 0.9845 & 0.7931  \\ \hline
		DCN & 1.3699 & 1.1509 & 0.9964 & 0.8105 & 0.9461 & 0.7575 & 0.9742 & 0.7863 \\ \hline
		DeepFM & 1.3781 & 1.1581 & 0.9840 & 0.7962 & 0.9693 & 0.7817 & 0.9827 & 0.7935 \\ \hline
		FiBiNET & 1.4415 & 1.2117 & 0.9922 & 0.7691 & 0.9723 & 0.7790 & 0.9551 & 0.7672 \\ \hline
		NFM & 1.3896 & 1.2143 & 0.8957 & 0.7124 & 0.9565 & 0.7664 & 0.9383 & 0.7561 \\ \hline
		ONN & 1.4389 & 1.2359 & 1.0071 & 0.8015 & 1.0403 & 0.8379 & 1.0203 & 0.8284 \\ \hline
		WDL & 1.3742 & 1.1570 & 0.9885 & 0.7992 & 0.9792 & 0.7892 & 0.9743 & 0.7911 \\ \hline
		xDeepFM & 1.4102 & 1.1571 & 0.9198 & 0.6970 & 0.9134 & 0.6891 & 0.9172 & 0.6960 \\ \hline
		MHA &\textbf{ 1.3657} & \textbf{1.1212} & \textbf{0.8798} & \textbf{0.6478} & \textbf{0.8823} & \textbf{0.6617} & \textbf{0.8484} & \textbf{0.6529} \\ \hline
	\end{tabular}
	
	\vspace{-0.6cm}
\end{table}

In general, our MHA architecture consistently demonstrates superior performance across all four scenarios compared to the baseline methods. For example, in the GRS scenario without context or criteria, MHA achieves the lowest RMSE (1.3657) and MAE (1.1212), outperforming methods such as FiBiNET and AutoInt, which show higher RMSE values of 1.4415 and 1.3765, respectively. This highlights the MHA architecture's capacity to better capture intricate group preferences and interactions, showcasing its ability to model complex relationships in recommendation tasks more effectively than other approaches. Furthermore, the architecture's robustness across varying conditions indicates its suitability for handling multi-criteria and context-aware group recommendations.

The comparison of the four scenarios shows that incorporating criteria and contextual factors significantly improves the performance of the recommendation model. Starting from the basic GRS without context and criteria, the model's accuracy increases as multi-criteria data is introduced, reducing RMSE and MAE. Further enhancements occur when multiple contextual factors are added, highlighting the importance of context in shaping group preferences. The best results are achieved when using MCGRS with a single contextual factor (class context), demonstrating that even focused contextual information can substantially improve recommendation quality. Overall, both criteria and context are crucial for achieving more accurate and relevant group recommendations.

\begin{figure}[h!]
	\vspace{-0.3cm}
	\centering
	\includegraphics[width=0.9\textwidth]{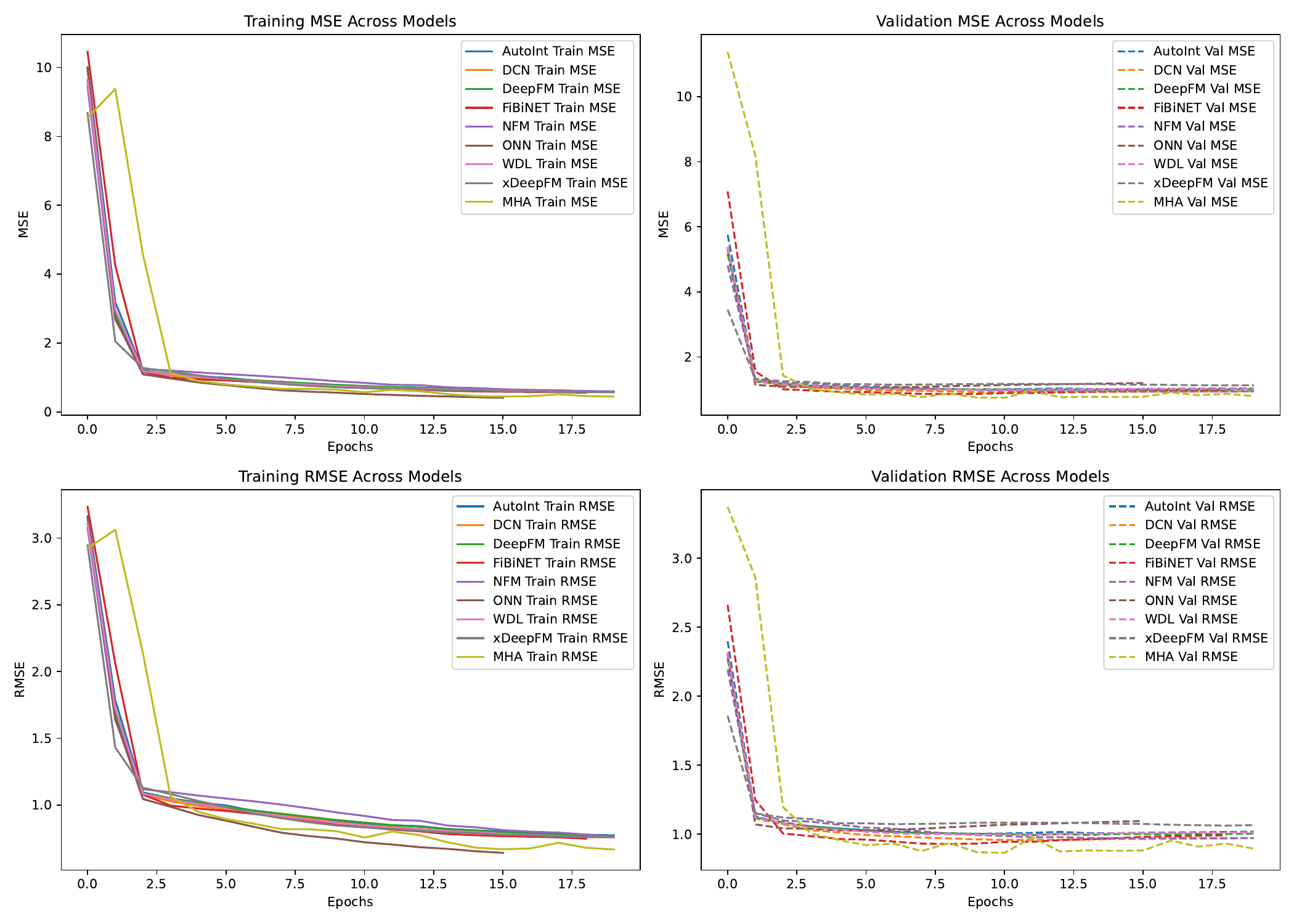}
	\vspace{-0.3cm}	
	\caption{Comparison of the performance of different models across training and validation sets. The top-left subplot shows the Training MSE for all models, while the top-right subplot depicts the Validation MSE. The bottom-left and bottom-right subplots illustrate the Training RMSE and Validation RMSE respectively, providing a view of each model's prediction error over the course of training epochs. Models with lower MSE and RMSE values demonstrate better accuracy in the recommendation task.} \label{fig2}
\end{figure}

As shown in figure \ref{fig2}, the training and validation curves reinforce these findings. MHA not only converges faster during training but also exhibits significantly lower errors in the validation phase. While models such as DeepFM and FiBiNET show competitive training performance, their validation errors remain higher, suggesting potential overfitting or difficulty in generalizing to unseen data. In contrast, MHA maintains low validation errors, highlighting its robustness and ability to generalize well.

The experimental results clearly demonstrate the effectiveness of the CA-MCGRS architecture in handling group-based recommendations, especially in multi-criteria and context-aware settings. Across all scenarios, including GRS without context, MCGRS without context, and MCGRS with both single and multiple contextual factors, the CA-MCGRS architecture consistently achieved the lowest RMSE and MAE values. This performance advantage highlights the model's ability to adaptively capture complex group preferences, context, and criteria interactions. The significant improvement in accuracy, particularly in context-aware scenarios, reinforces the strength of our architecture in addressing the nuanced requirements of GRS. These results underscore the potential of CA-MCGRS architecture to enhance the precision and relevance of recommendations in a variety of real-world applications.


\section{Conclusion}
In this paper, we explored the development of a context-aware multi-criteria group recommender system (CA-MCGRS) using a multi-head attention mechanism (MHA). Our approach dynamically identifies and weighs features from various contexts and criteria, allowing the model to adapt to different recommendation scenarios. By utilizing multiple attention heads, our model captures intricate relationships between groups, items, and contextual factors, making it highly effective in complex recommendation tasks. We evaluated the MHA-based model against several leading baselines using RMSE and MAE as performance metrics. The results demonstrate that our approach consistently outperforms traditional methods, achieving superior accuracy and lower error rates across all scenarios. This improvement is due to the MHA's capacity to effectively prioritize and integrate relevant features dynamically. These findings highlight the potential of attention-based mechanisms in advancing RSs by providing more precise and context-aware recommendations.

 Future work will focus on further integrating complex contextual information, improving model scalability, and increasing the interpretability of deep learning models in CA-MCGRS. Additionally, testing and scaling the CA-MCGRS on other datasets will be critical in assessing its adaptability and performance in different domains. Evaluating the model on larger and more complex datasets will help determine its efficiency, generalization capabilities, and ability to handle data sparsity effectively.
 \bibliographystyle{splncs04}
 \bibliography{references}
\end{document}